%% file: main.tex
\documentclass[10pt,twocolumn,letterpaper]{article}

\usepackage[pagenumbers]{iccv} 

\input{preamble}

\definecolor{iccvblue}{rgb}{0.21,0.49,0.74}
\usepackage[pagebackref,breaklinks,colorlinks,allcolors=iccvblue]{hyperref}

\def\paperID{8607} 
\def\confName{ICCV}
\def\confYear{2025}

\title{Puzzle Similarity: A Perceptually-guided Cross-Reference Metric for Artifact Detection in 3D Scene Reconstructions}

\author{Nicolai Hermann\textsuperscript{1,2} \qquad Jorge Condor\textsuperscript{1,2} \qquad Piotr Didyk\textsuperscript{1,2}\\\\\textsuperscript{1}USI, Lugano, Switzerland\\\textsuperscript{2}IDSIA, Switzerland\\\{nicolai.hermann, jorge.condor, piotr.didyk\}@usi.ch
}

\begin{document}
\maketitle
\input{sec/00_definitions}

\input{sec/0_abstract}
\input{sec/1_intro}
\input{sec/2_related_work}
\input{sec/3_methods}

\input{sec/4_results}
\input{sec/5_application}
\input{sec/6_limitations}
\input{sec/7_conclusion}
{
    \small

\input{main.bbl}
}
\end{document}

%% file: preamble.tex
\usepackage{booktabs}
\usepackage{longtable}
\usepackage{multirow}
\usepackage{xcolor}
\usepackage{float}
\usepackage{tabularx}
\usepackage{adjustbox}
\usepackage{array}
\usepackage{booktabs}
\usepackage{colortbl}
\usepackage{graphicx}
\usepackage{caption}
\usepackage{subcaption}
\usepackage{soul}
\usepackage{arydshln}
\usepackage[accsupp]{axessibility}  
\newcommand{\reduce}{\vspace{-14pt}}

\newcommand{\revision}[2]{{\small\st{#1}}{\color{Bittersweet}#2}}
\newcommand{\finalrev}[1]{{\color{Bittersweet}#1}}
\renewcommand{\revision}[2]{#2}
\renewcommand{\finalrev}[1]{#1}

\definecolor{cvprblue}{rgb}{0.21,0.49,0.74}
\definecolor{lightred}{rgb}{1,0.4,0.4}
\definecolor{lightyellow}{rgb}{1,1,0.4}
\definecolor{lightorange}{rgb}{1,0.647,0.4}

%% file: sec/00_definitions.tex
\newcommand{\img}{\mathcal{I}}
\newcommand{\R}{\mathbb{R}}

\newcolumntype{R}[1]{%
    >{\adjustbox{angle=#1,clap}\bgroup}%
    c%
    <{\egroup}%
}
\newcolumntype{F}[2]{%
    >{\adjustbox{angle=#1,lap=\width-(#2)}\bgroup}%
    l%
    <{\egroup}%
}
\newcommand*\rot{\multicolumn{1}{R{45}}}
\newcommand*\rotw{\multicolumn{1}{F{60}{1em}}}
\newcommand{\1}{\cellcolor{lightred}}
\newcommand{\2}{\cellcolor{lightorange}}
\newcommand{\3}{\cellcolor{lightyellow}}

%% file: sec/0_abstract.tex
\begin{abstract}
Modern reconstruction techniques can effectively model complex 3D scenes from sparse 2D views. However, automatically assessing the quality of novel views and identifying artifacts is challenging due to the lack of ground truth images and the limitations of no-reference image metrics in predicting reliable artifact maps. The absence of such metrics hinders assessment of the quality of novel views and limits the adoption of post-processing techniques, such as inpainting, to enhance reconstruction quality. To tackle this, recent work has established a new category of metrics (cross-reference), predicting image quality solely by leveraging context from alternate viewpoint captures \cite{wang_crossscore_2025}. In this work, we propose a new cross-reference metric, Puzzle Similarity, which is designed to localize artifacts in novel views. Our approach utilizes image patch statistics from the training views to establish a scene-specific distribution, later used to identify poorly reconstructed regions in the novel views. Given the lack of good measures to evaluate cross-reference methods in the context of 3D reconstruction, we collected a novel human-labeled dataset of artifact and distortion maps in unseen reconstructed views. Through this dataset, we demonstrate that our method achieves state-of-the-art localization of artifacts in novel views, correlating with human assessment, even without aligned references. We can leverage our new metric to enhance applications like automatic image restoration, guided acquisition, or 3D reconstruction from sparse inputs. Find the project page at \url{https://nihermann.github.io/puzzlesim/}.
\end{abstract}

%% file: sec/1_intro.tex
\section{Introduction}\label{sec:introduction}
Image-based rendering and 3D reconstruction from a sparse set of 2D views has received ample attention in recent years, both for pure geometry reconstruction and radiance-field modeling. \finalrev{Classical approaches such as structure from motion (SfM) use simple triangulation and epipolar geometry to produce sparse point clouds of diffuse color~\cite{schonberger_structure--motion_2016}.} Densifying these representations can be done explicitly~\cite{kerbl_3d_2023}. Alternatively, one can learn continuous, implicit representations~\cite{mildenhall_nerf_2020, barron_mip-nerf_2021, muller_instant_2022}, normally modeled through multi-layer perceptrons.
A tangential problem to these efforts is the collection of 2D data and the handling of corrupted, distorted, or simply incomplete sets of images from an object or scene we would like to model. \finalrev{Learning representations from very sparse inputs has been a widely studied topic, where normally learned priors from large datasets are leveraged to enforce 3D consistency to ensure that the resulting reconstructions follow natural statistics \cite{yu_pixelnerf_2021, chen_fast_2024, warburg_nerfbusters_2023, chung_depth-regularized_2024}.}
However, quantifying the quality of novel views from reconstructions is still problematic. These views can contain artifacts due to the sparsity of the training dataset, and automatically identifying them helps with restoration (e.g., masking for image-based inpainters~\cite{suvorov_resolution-robust_2022}) or simply to guide future data acquisition to fill the gaps~\cite{kopanas_improving_2023}. Recent works have followed a Bayesian approach to quantify the uncertainty of whether an area belongs to a reconstructed scene or not~\cite{goli_bayes_2023}, which could potentially be leveraged for simple artifact detection. However, they require implicit models, with fundamental changes to the scene model, and are not practical for more general applications that require visual artifact identification outside of scene reconstruction, and are incapable of detecting artifacts not arising from lack of coverage. 

To tackle this, we propose a novel approach for artifact detection that can be leveraged on any set of images without an encoded explicit or implicit model of the scene or object they depict. Unlike visual difference predictors (VDPs)~\cite{mantiuk_visible_2004} (which require references) and no-reference quality metrics~\cite{mittal_no-reference_2012, mittal_making_2013} (which typically do not provide maps, but rather produce single values of overall quality), our approach provides visual artifact maps with no direct references. We leverage learned perceptual patch statistics from small, clean datasets and compare them to the embedded statistics of new images from a similar distribution (i.e., novel reconstructed views from a 3DGS~\cite{kerbl_3d_2023} representation) to obtain artifact maps without aligned references. We test our generated maps through a human experiment where we ask participants to manually identify artifacts and distortions in images to generate ground-truth data of visual artifacts. Our results show that our method agrees with human assessment, correlating better than no-reference, full-reference, and state-of-the-art cross-reference metrics.
To summarize, our contributions are the following:
\begin{itemize}
    \item A novel cross-reference visual artifact identification metric, particularly tailored for 3D reconstruction,
    \item a novel dataset of human-labeled artifact and distortion maps to fill the gap of validation benchmarks for cross-reference metrics,
    \item and applications on image restoration and 3D reconstruction enhancement that showcase our approach's utility.
\end{itemize}

%% file: sec/2_related_work.tex
\section{Related Work}\label{sec:related_work}
Our metric is specifically designed for applications in 3D scene reconstruction and image-based rendering. Consequently, this section discusses work on 3D reconstruction first and then on image metrics.

\subsection{3D Reconstruction and Image-based Rendering}
Reconstructing 3D objects or scenes from sparse sets of 2D observations is a fundamental problem in vision~\cite{mantiuk_visible_2004}. Particularly, in the context of novel view synthesis, the objective is to approximate the radiance field (i.e., 5D function encoding spatially varying radiance emission) of specific objects or scenes. Most methods differ either in the model used to encode the function or the rendering procedure.
Implicit approaches model the radiance field as a continuous function, approximated by a multi-layer perceptron~\cite{mildenhall_nerf_2020, sitzmann_deepvoxels_2019}. Rendering is usually done via sampling the implicit volume using ray-marching~\cite{tuy_direct_1984}, which provides spatially varying values of density and anisotropic color emission modeled through Spherical Harmonics. Improvements over this formula have tackled performance limitations, either by using more efficient sampling techniques~\cite{neff_donerf_2021, gupta_mcnerf_2023, muller_instant_2022} or by distilling the implicit space into explicit density and anisotropic appearance volumes~\cite{yu_plenoctrees_2021, xu_wavenerf_2023}. 
On the other hand, purely explicit models do not require any pre-training using implicit functions, and were originally Eulerian in nature~\cite{yu_plenoxels_2021}. Explicit models are easier to optimize, usually faster, and more interpretable, which can help in different tasks such as scene editing or animation. More recently, anisotropic Lagrangian approaches have found tremendous success~\cite{kerbl_3d_2023}.
However, these explicit methods have introduced some limitations of their own along the way. Methods like 3D Gaussian Splatting~\cite{kerbl_3d_2023} can only model areas that are directly supervised, and degrade less gracefully than implicit counterparts when querying viewpoints substantially outside the training set coverage. Detecting artifacts arising from the lack of coverage is difficult due to the lack of reference images. Our method produces these masks via supervision on the training data solely, which can enable unsupervised restoration (automatic inpainting of the artifacts based on available context~\cite{fawzi_image_2016}) or simply automatically guide further image acquisition to complete the dataset efficiently~\cite{kopanas_improving_2023}.

\subsection{Image Metrics}
\finalrev{Image metrics are classified by the type of prediction they make and their input. Image Quality Metrics (IQMs) typically predict a single number, corresponding to overall image quality, and are often trained on Mean Opinion Score (MOS) datasets. Visibility Metrics (VMs) produce maps corresponding to the perceptibility of distortions. They often rely on models of the human visual system and predict the probability of detecting local artifacts by an observer. Most image metrics are full-reference, necessitating a reference to assess the quality of a test image. In contrast, no-reference metrics predict the quality or visible distortions based solely on the test image. Others use additional information, e.g., partial reference, and are referred to as reduced-reference or cross-reference metrics.

Classical examples of full-reference IQMs include mean-absolute error (MAE), mean-squared error (MSE), peak signal-to-noise ratio (PSNR), SSIM~\cite{wang_image_2004}, FSIM~\cite{zhang_fsim_2011}, MS-SSIM~\cite{wang_multiscale_2003}, and LPIPS~\cite{zhang_unreasonable_2018}. These methods begin by computing local differences between test and reference images and aggregate them into a single quality score as the final step. By omitting this step, it is possible to create a local distortion map, which is a common output of VMs widely adopted in rendering to localize poorly rendered areas \cite{andersson_flip_2020}. Typically, VMs differ from IQMs in their more explicit modeling of the human visual system \cite{daly_visible_1992}, which enables the prediction of perceived visibility or the strength of visual differences between images. Improvements over the original framework extended their applicability to high dynamic range imagery \cite{mantiuk_hdr-vdp-3_2023}, making them eccentricity and motion-aware \cite{tursun_perceptual_2022,mantiuk_fovvideovdp_2021}, and integrating perceived color \cite{mantiuk_colorvideovdp_2024}. Such metrics have been used in many perceptual optimizations, such as foveated rendering \cite{tursun_luminance-contrast-aware_2019} and perceptually aware tone mappers \cite{tariq_perceptually_2023}.

No-reference metrics eliminate the need for a reference and are most commonly learned from human quality assessment datasets \cite{ke_musiq_2021,kang_convolutional_2014,ying_patches_2020}, supervised on extracted features from natural image statistics \cite{moorthy_blind_2011, mittal_no-reference_2012, xue_learning_2013, mittal_making_2013}, or even synthetic scores \cite{ye_beyond_2014}.
Modern deep learning approaches can utilize Transformer architectures \cite{you_transformer_2021}, and multi-scale Transformers are employed to alleviate the resolution constraints \cite{ke_musiq_2021}. Hybrid approaches use multi-modal architectures in conjunction with text templates to query image features such as noisiness, sharpness, or contrast, which can be translated into MOS \cite{zhang_blind_2023, wang_exploring_2023}. The above no-reference metrics rely on global image features and, therefore, are not suitable for obtaining distortion maps. However, similarly to our work, some of these IQMs are capable of producing visual maps. For example, PIQE \cite{n_blind_2015} measures distortions in an image patch based on extracted local features. CNNIQA~\cite{kang_convolutional_2014}, on the other hand, was one of the first no-reference metrics to employ convolutional neural networks (CNN) to regress mean-opinion scores. More recently, PaQ-2-PiQ~\cite{ying_patches_2020} uses region proposals to select quality-determining patches. PAL4VST~\cite{zhang_perceptual_2023} localizes specific artifacts that emerge from image synthesis tasks through binary segmentation. In contrast, our work leverages the latent space of a model pre-trained on natural images to measure the cosine similarity in feature space of candidate image patches to a limited set of image patches from a similar distribution (i.e., images from the same scene in the context of scene reconstruction), rendering a higher level of alignment with human assessment.

In some applications, even though an image metric does not have access to a reference image, it may have access to other information useful for making a prediction. In the context of novel view synthesis, cross-reference metrics utilize a set of unaligned reference images to assess the quality of a single query image belonging to the same scene. This type of metric was recently established by Wang et al.~\cite{wang_crossscore_2025} and their metric CrossScore serves as the main baseline for our evaluation. The metric relies on a cross-attention module~\cite{vaswani_attention_2017} to correlate a test image with unaligned multiview images to predict quality maps. The maps predict the quality of 14x14 patches of the input image and are trained to mimic unpooled SSIM maps. However, SSIM has been repeatedly shown to be poorly aligned with human quality assessment and perception \cite{nilsson_understanding_2020, pambrun_limitations_2015, zhang_unreasonable_2018}, which fundamentally limits the potential of CrossScore.

In comparison to previous work, our method is a cross-reference metric for novel view synthesis tasks. It predicts local similarities for a synthesized view given a set of unaligned reference views. Its effectiveness in localizing artifacts is achieved by assessing similarity in feature space. In contrast to many IQMs, our method is not designed to predict a single score, but rather a map corresponding to the strength of the visible artifacts.
}

%% file: sec/3_methods.tex
\section{Our Method}\label{sec:methods}
\begin{figure*}[ht!]
    \centering
    \includegraphics[width=1\textwidth]{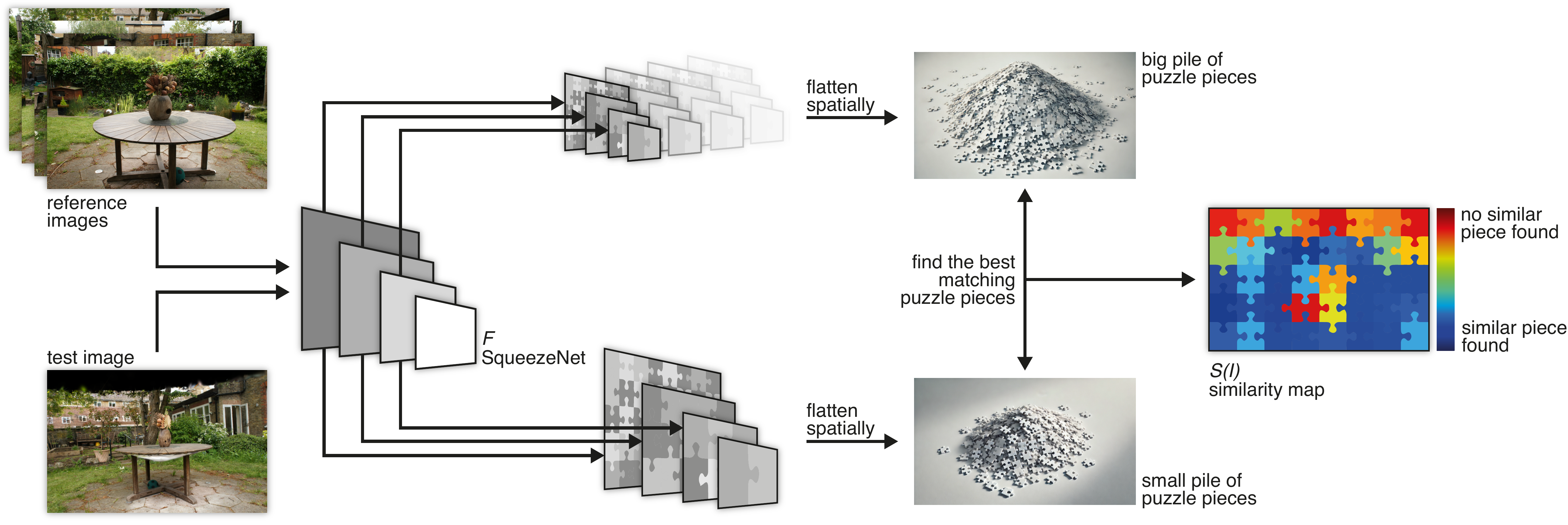}
    \caption[Schematic representation of our Puzzle Similarity metric]{Schematic representation of our Puzzle Similarity metric.}
    \label{fig:puzzle-schematic}
\end{figure*}
Let us establish an analogy for our method: pretend each reference image is a puzzle with many puzzle pieces. To test if a new image is similar to our unaligned references, we would simply shuffle all pieces from all puzzles from our references and try to reassemble the test image only using those pieces. If the new image is very similar to the references, we should have enough puzzle pieces to compose the other image confidently. However, if the image holds regions very different from what we saw in the reference images, we would lack puzzle pieces to assemble this area, effectively leaving holes in the newly assembled puzzle (image). An overview of our approach through this analogy can be seen in \cref{fig:puzzle-schematic}.

\begin{figure}
    \centering
    \includegraphics[width=.5\linewidth]{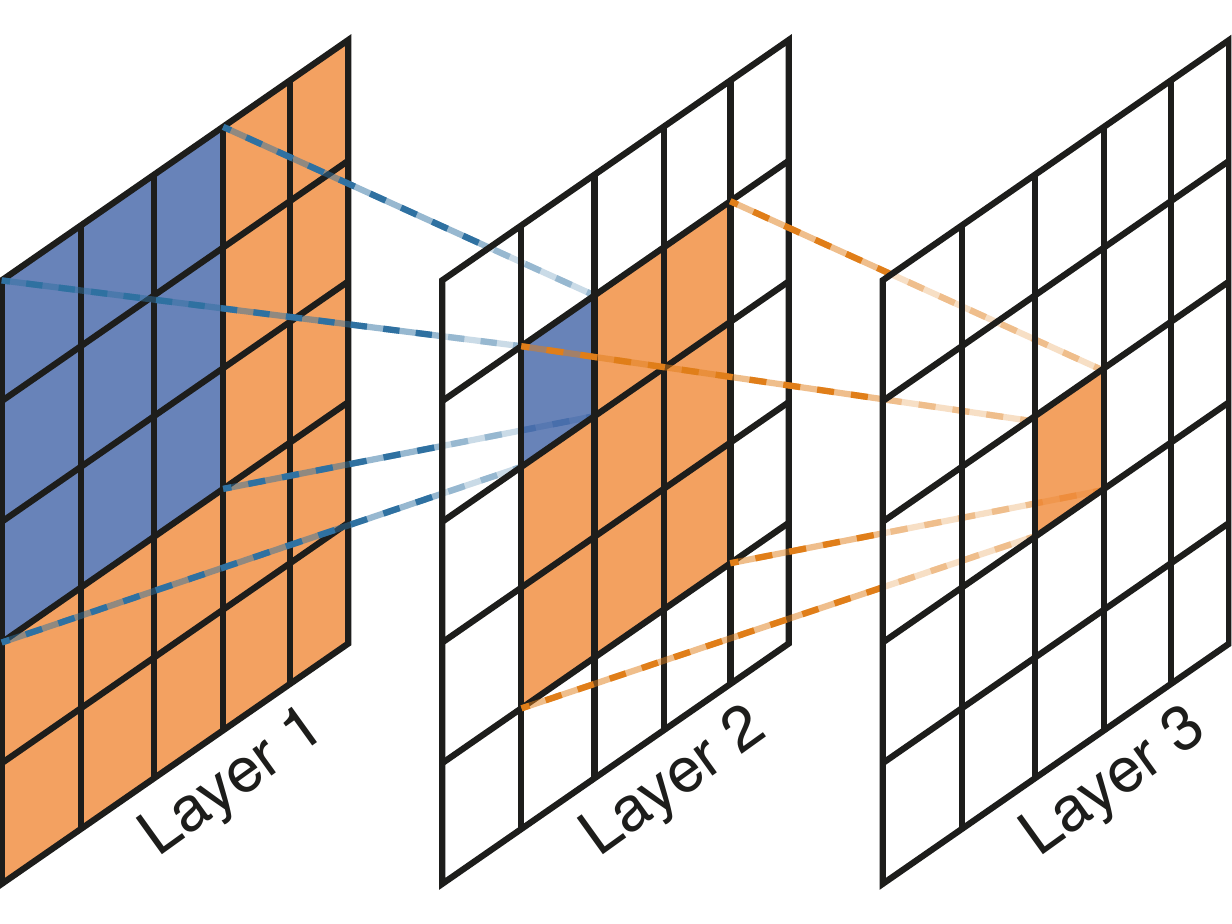}
    \caption{Receptive field of a multi-layer CNN. Note how one pixel in the last layer is an embedding of a patch of the input space.}
    \label{fig:receptive_field}
\end{figure}

In our work, the puzzle pieces correspond to embedded image patches. In order to assess patch similarity, an obvious approach would involve computing the dot product between all patches; best-matching pieces would be recorded to create a similarity map. This simplistic approach, however, would hardly align with human assessment if unprocessed patches were used. Inspired by the close correlation between human quality judgment and latent CNN feature maps~\cite{vedaldi_why_2020, zhang_unreasonable_2018}, we employ a pre-trained CNN~\cite{krizhevsky_imagenet_2012, iandola_squeezenet_2016, simonyan_very_2015} to embed all the references, computing similarity in the latent feature space. Note that comparing feature map "pixels" in a CNN is similar to comparing individual patches in the input domain; this is due to the locality of the sliding kernels when convolving. The patch size is dependent on the receptive field (showcased in \cref{fig:receptive_field}).

\reduce
\paragraph{Choice of layers}
Choosing the right layers for embedding is essential to maximize the quality of the predicted spatial maps. While early layers feature small receptive fields and capture fine details, deeper layers have larger receptive fields and capture coarser features. This can be observed in \cref{fig:lpips_layers}, where we showcase different \verb|VGG| layers. It is essentially a trade-off between prediction granularity, accuracy and speed. We identified that combining multiple layers into our metric computation incorporates the various levels of abstraction and scales in a robust manner. We thus compute the weighted average of the three layers; we empirically found that halving the image resolution more than three times did not significantly improve our results, as the scale becomes too small and the pool of reference vectors too little and specific to find good correspondences among novel images, even for well-reconstructed areas. This observation suggested that features from the layers before the third down-sampling were most useful for our cause. 

\reduce
\paragraph{Computing patch similarity}
To compute the similarity map of a test image, we feed all references and the test image through a pre-trained network $\mathcal{F}$ to obtain the embeddings. We repeat the exact computation for each network layer; thus, we will describe the steps once for one layer $\ell$. To find the similarity $s_\ell(x, y)$ to the best matching puzzle piece for a pixel of the embedded test image at some spatial location $(x, y)$, we compute the cosine similarity based on the feature vector $\mathcal{F}_\ell(x, y)$ and all other feature vectors of all $N$ reference images of the same layer $\ell$ and select the correspondence with the highest similarity:
\begin{equation}
    \begin{aligned}
        s_\ell(x, y) = \max_{n, x', y'} \hat{\mathcal{F}}_\ell(x, y) \cdot \hat{\mathcal{F}}^{(n)}_\ell(x', y')
    \end{aligned}
\end{equation}
where $\hat{\mathcal{F}}$ denotes a feature vectors scaled to unit length $\hat{\mathcal{F}}_\ell(x, y) = \frac{\mathcal{F}_\ell(x, y)}{||\mathcal{F}_\ell(x, y)||_2} \in \R^{C_\ell}$ and $\cdot$ is the dot product. Note that we compute the cosine similarity with any feature vector of the same layer from all references, not just those at the same spatial position. This relinquishes spatial relations and makes the method robust to simple camera movements that only shift the image horizontally or vertically. We iterate this maximum search for all pixels of the test image's embedding to construct the similarity mask $\mathcal{S}_\ell$.
\begin{equation}
    \begin{aligned}
        \mathcal{S}_\ell(\img) = 
\begin{bmatrix}
s_\ell(1, 1) & s_\ell(1, 2) & \cdots & s_\ell(1, W_\ell) \\
s_\ell(2, 1) & s_\ell(2, 2) & \cdots & s_\ell(2, W_\ell) \\
\vdots & \vdots & \ddots & \vdots \\
s_\ell(H_\ell, 1) & s_\ell(H_\ell, 2) & \cdots & s_\ell(H_\ell, W_\ell) \\
\end{bmatrix}
    \end{aligned}
\end{equation}
where $\img$ is the test image. We repeat this for a set of layers and combine them into a final similarity map. To match the spatial dimensions of each layer, we bilinearly upsample each map to the original image size and combine them with an affine combination:
\begin{equation}
    \begin{aligned}
        \mathcal{S}(\img) &= \sum_{\ell} w_\ell \; \text{Upsample}\left(\mathcal{S}_\ell(\img, \; \img_{\text{ref}}^{1:N})\right)
    \end{aligned}
\end{equation}
with $\sum_\ell w_\ell = 1$ and reference images $\img_{\text{ref}}^{1:N}$. To utilize optimized hardware, please note how the computation of $S_\ell$ can also be expressed as an outer product between the spatially flattened embeddings:
\begin{equation}
    \begin{aligned}
    \hat{\mathcal{F}}_\ell(\img^{1:N}) &\in \R^{N \times H_\ell \times W_\ell \times C_\ell} \\
        \tilde{\mathcal{F}}_\ell(\img^{1:N}) &= \text{flatten}\left(\hat{\mathcal{F}}_\ell(\img^{1:N})\right) \in \R^{NH_\ell W_\ell \times C_\ell} \\
        \mathcal{S}_\ell(\img) &= \overbrace{ \text{ rowmax } \underbrace{\tilde{\mathcal{F}}_\ell(\img_{\text{ref}}^{1:N}) \otimes  \tilde{\mathcal{F}}_\ell(\img)}_{\in \; \R^{NH_\ell W_\ell \times H_\ell W_\ell}} }^{\in \; \R^{H_\ell W_\ell}}
    \end{aligned}
\end{equation}
where the test image $\img$ is a special case with $N=1$, $\otimes$ is the outer product, and $\text{rowmax}$ applies the max over the first dimension. While a naïve implementation of this outer product would require substantial amounts of memory for larger $N, H, W$, we provide an efficient implementation through blockwise tiling with intermediate max-reduction, which we detail in the Supplemental. 
\input{figures/vgg_layer_maps}
\reduce
\paragraph{Pre-trained Model Choice}
The choice of pre-trained neural network, through which the embeddings will be created, is a key component of our work. We primarily considered classic models including \verb|VGG-16|, \verb|VGG-19|, \verb|AlexNet|, and \verb|SqueezeNet|~\cite{simonyan_very_2015, krizhevsky_imagenet_2012, iandola_squeezenet_2016}. Some of the critical considerations are model complexity and memory requirements, which we summarized in the Supplemental, as well as their specifically tested performance on our human alignment task. Beyond quality performance, reducing the memory footprint and computational complexity is key as it may impact the possibility of downstream applications of our metric, which, given its differentiability, could be leveraged in optimization procedures.

We empirically found that while \verb|VGG| produces the most fine-grained maps, \verb|AlexNet| and \verb|SqueezeNet| still managed to perform similarly, while doing so at a substantially reduced computational cost. We opted for \verb|SqueezeNet| as it aligned best with our test examples, specifically using layers $\ell \in \{2, 3, 4\}$ with the weights $w_2 = 0.67$, $w_3 = 0.2$, and $w_4 = 0.13$, which we found heuristically.

%% file: figures/vgg_layer_maps.tex
\begin{figure*}
    \centering
    \begin{subfigure}[b]{0.24\textwidth}
        \centering
        \includegraphics[width=\textwidth]{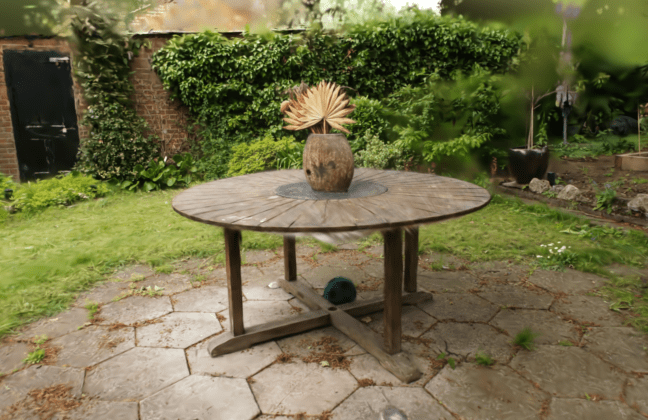}
        \caption{Rendering}
    \end{subfigure}
    \hfill
    \begin{subfigure}[b]{0.24\textwidth}
        \centering
        \includegraphics[width=\textwidth]{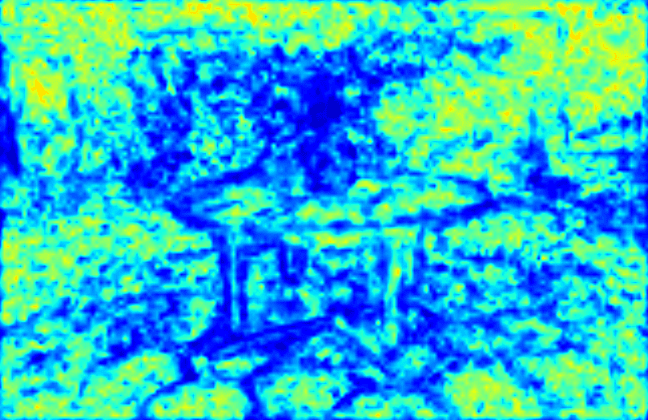}
        \caption{$2^{\text{nd}}$ VGG layer}
    \end{subfigure}
    \hfill
    \begin{subfigure}[b]{0.24\textwidth}
        \centering
        \includegraphics[width=\textwidth]{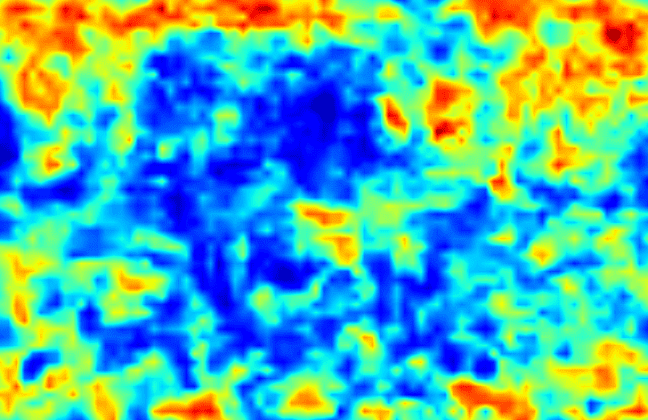}
        \caption{$3^{\text{rd}}$ VGG layer}
    \end{subfigure}
    \hfill
    \begin{subfigure}[b]{0.24\textwidth}
        \centering
        \includegraphics[width=\textwidth]{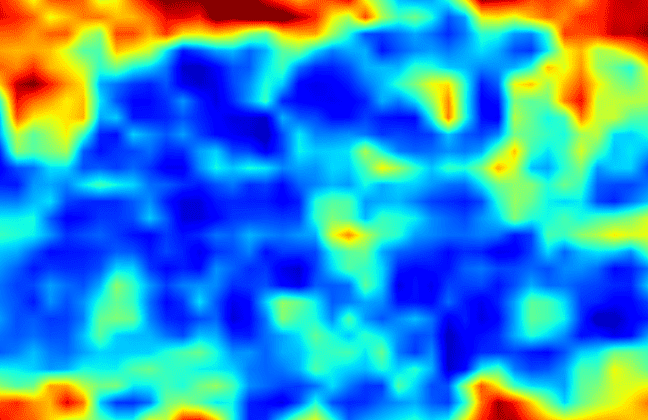}
        \caption{$4^{\text{th}}$ VGG layer}
    \end{subfigure}
    
    \caption[Analysis of the layers of our Puzzle Similarity metric]{Puzzle Similarity computed on a single VGG layer. Note how the second layer has a finer resolution and mostly cold colors, while the fourth layer is much smoother and features a wider range of values. Warm colors indicate artifacts or poor reconstruction quality.}
    \label{fig:lpips_layers}
\end{figure*}

%% file: sec/4_results.tex
\section{Results}\label{sec:results}

\input{figures/maps_figure}
We will now analyze how our method compares against competing approaches for both no-reference and cross-reference visual map prediction in the context of reconstruction and image-based rendering. 
To quantify the correlation between all these maps and human assessment, we present a novel dataset on human artifact identification, which we manually collected and can be found here\footnote{\scriptsize\url{https://huggingface.co/datasets/nihermann/annotated-3DGS-artifacts}} to facilitate future research on the topic.
As for our method, for each different scene, we compute embeddings on their respective training dataset to compute similarity maps on test views, as explained in \cref{sec:methods}. CrossScore leverages the same set of training views for its map predictions.

\subsection{A Novel Artifact Identification Dataset}\label{sec:dataset}
We created a dataset of human-perceived artifacts in 3D reconstructed views with corresponding ground truths collected through a user study. \revision{}{To generate images exhibiting typical reconstruction artifacts, we apply 3D Gaussian Splatting~\cite{kerbl_3d_2023} to twelve scenes from the Mip-NeRF360~\cite{barron_mip-nerf_2021}, Tanks and Temples~\cite{knapitsch_tanks_2017}, and Deep Blending~\cite{hedman_deep_2018} datasets. We use default parameters but intentionally withhold a significant portion of training images, which we later utilize for additional validation. By omitting these views during training, we increase the likelihood of artifacts appearing in the withheld views.} For each dataset, we selected three renderings that demonstrated a mix of well-reconstructed areas, strong artifacts, and subtle artifacts, resulting in 36 samples across 12 datasets.

\reduce
\paragraph{Experiment details}
We asked 22 participants to segment visible artifacts in each of the 36 sample images under controlled viewing conditions using the tool developed by Wolski et al.~\cite{wolski_dataset_2018}, which the authors kindly provided. We include details on the participants' self-reported gender and age distributions in the Supplementary, as well as detailed viewing and display conditions. During the experiment, users had no undistorted, artifact-free references at their disposal and thus had to judge individual images at face value. They would then mark areas found to be unnatural or unappealing, creating a binary mask. 
With the dataset, we can evaluate the agreement between human judgment and any metric output by simply averaging all binary masks to estimate the probability of each pixel being marked as an artifact. \cref{fig:examples-experiment} shows example renderings (a) alongside metric predictions (b)-(d) and their average human-produced mask (e). 

We evaluate our method against both no-reference and \revision{}{cross-reference} metrics. To assess their alignment with human perception, we correlate their maps with the human ratings from our dataset, as described in \cref{sec:dataset}. \revision{}{NR and CR metrics are the only metrics that can detect artifacts without a direct reference, but the way we collected our dataset gives us access to a ground truth that is normally unavailable. This enables us to assess FR metrics and current VDPs too, although they are otherwise not suitable for the objective in question. We include extensive results in the Supplemental and show that, unlike any other competing metric, our metric even outperforms the best FR metrics on this benchmark, proving the general superiority of our method.}

\subsection{Evaluation}
\input{tables/no_ref_table}

To correlate metric outputs to our human segmentations, we first compute each metric map for each rendering and then compute the Pearson correlation coefficient (PCC) and Spearman's rank correlation coefficient (SRCC). To account for the different domains of the compared metrics and possibly non-linear relations, we fit a 5-parameter logistic curve for a fair comparison as suggested by \cite{adhikarla_towards_2017, li_deep_2024, zhang_feature-enriched_2015, sheikh_statistical_2006}:
\begin{equation}
    \begin{aligned}
        q(x) = a_1 \left\{\frac{1}{2} - \frac{1}{1 + \exp{(a_2 (x - a_3))}} \right\} a_4 x + a_5
    \end{aligned}
\end{equation}
where $x$ is an individual pixel score and $a_{1\dots5}$ are optimized through gradient ascent to maximize PCC or SRCC.

\vspace{2pt}
\reduce
\paragraph{Results}
\cref{tab:nr_per_ds} reports the average Pearson and Spearman correlations for each scene. Our dataset includes three images per scene, with varying artifact types per scene. For example, the \textit{garden} scene has prominent black regions due to holes in the reconstruction, making artifact detection straightforward and leading to high correlation scores for most metrics. However, scenes like \textit{treehill}, \textit{stump}, and \textit{flowers} exhibit artifacts in the form of blurry or unnatural textures while preserving similar color distributions to the ground truth. Puzzle Similarity consistently achieves high correlation with human-perceived artifacts across all datasets, \revision{}{retaining smaller variance across datasets than all competitive methods (See \cref{tab:corr}). The higher variance in the averaged results is mainly due to scenes having different types of artifacts, where some are harder to identify than others (e.g., it is easier to identify black areas than small diffuse blobs). Keeping a small variance implies robust performance across various artifact types.} We include extended quantitative and qualitative results in our Supplemental.

\subsection{Comparison with No-Reference Metrics}
We compare with other no-reference metrics capable of producing spatial visibility maps~\cite{zagoruyko_learning_2015, kang_convolutional_2014, zhang_perceptual_2023, ying_patches_2020, wang_crossscore_2025}. CNNIQA~\cite{kang_convolutional_2014} was applied on patches as described in their paper. We applied padding to avoid cropping the borders and bilinearly upsampled the final map. PIQE~\cite{n_blind_2015} already produces three different kinds of maps that we averaged. PAL4VST~\cite{zhang_perceptual_2023}, PaQ-2-PiQ~\cite{ying_patches_2020} produce maps and were applied as described in their papers, but bilinearly upsampled to match the human maps' resolution.

Puzzle Similarity demonstrates superior accuracy in artifact localization, as shown by the correlation values in \cref{tab:nr_per_ds}. PAL4VST and CNNIQA performed poorly, as expected, given their focus on detecting specific types of distortions that are not necessarily similar to reconstruction artifacts. While PIQE and PaQ-2-PiQ performed well in certain scenes, their overall correlation with human opinion was generally lower in others, reflecting a less robust alignment with human assessment. However, while our method relies on a small subset of images from a similar distribution to the target image (e.g., the training dataset on novel view synthesis of a specific scene), NR metrics do not require any extra images and attempt to generalize to any input. 
\subsection{Comparison with Cross-Reference Metrics}
CrossScore is, to our knowledge, the only other CR metric besides ours that is also reliant on the training views. We also bilinearly upsampled its output to match the human maps' resolution. We show comparisons to CrossScore in \cref{tab:nr_per_ds} and \cref{tab:corr}, in \cref{fig:examples-experiment} and extended results in the Supplementary. We outperform CrossScore on most datasets and show better performance both on average and in terms of consistency (with a substantially smaller standard deviation among results). While the expensive domain-specific pretraining of CrossScore should, in theory, be superior to our feature-space patch matching leveraging general models pre-trained on collections of natural images, their reliance on SSIM as its target quality assessment metric limits its potential to accurately model human quality assessment, due to the well-known limitations of the metric in this regard~\cite{nilsson_understanding_2020, pambrun_limitations_2015, zhang_unreasonable_2018}. Their DINOv2 encoder limits map resolution to 14x14 blocks, reducing artifact localization fidelity. Furthermore, our approach is notably simpler: we can leverage any CNN as a feature encoder, allowing seamless adaptation to specific domains simply by swapping out the backbone. No retraining or distillation of any framework component is required.

\subsection{Comparison with Full-Reference Metrics}
Although full-reference metrics, unlike our method, require a direct reference image for detecting artifacts, we also provide an extensive comparison to them based on our dataset, which includes reference images. On average, we outperform all FR metrics, while CrossScore falls behind FovVideoVDP. Thus, in the context of 3D reconstruction, our metric performs better than all tested NR, CR, and FR metrics. Investigating why even the most advanced VDPs, rooted in complex models of low-level human visual processing, quantitatively fall behind remains a fascinating avenue for future work.

\input{tables/corr_new}

%% file: figures/maps_figure.tex
{
\renewcommand{\arraystretch}{0.5} 
\begin{figure*}[ht]
    \centering
    \begin{adjustbox}{width=0.96\textwidth,valign=t}
    \hspace{-0.024\textwidth}
        \begin{tabular}{c@{\hspace{0.1cm}}c@{\hspace{0.1cm}}c@{\hspace{0.1cm}}c@{\hspace{0.1cm}}c}
            \includegraphics[width=0.181\textwidth]{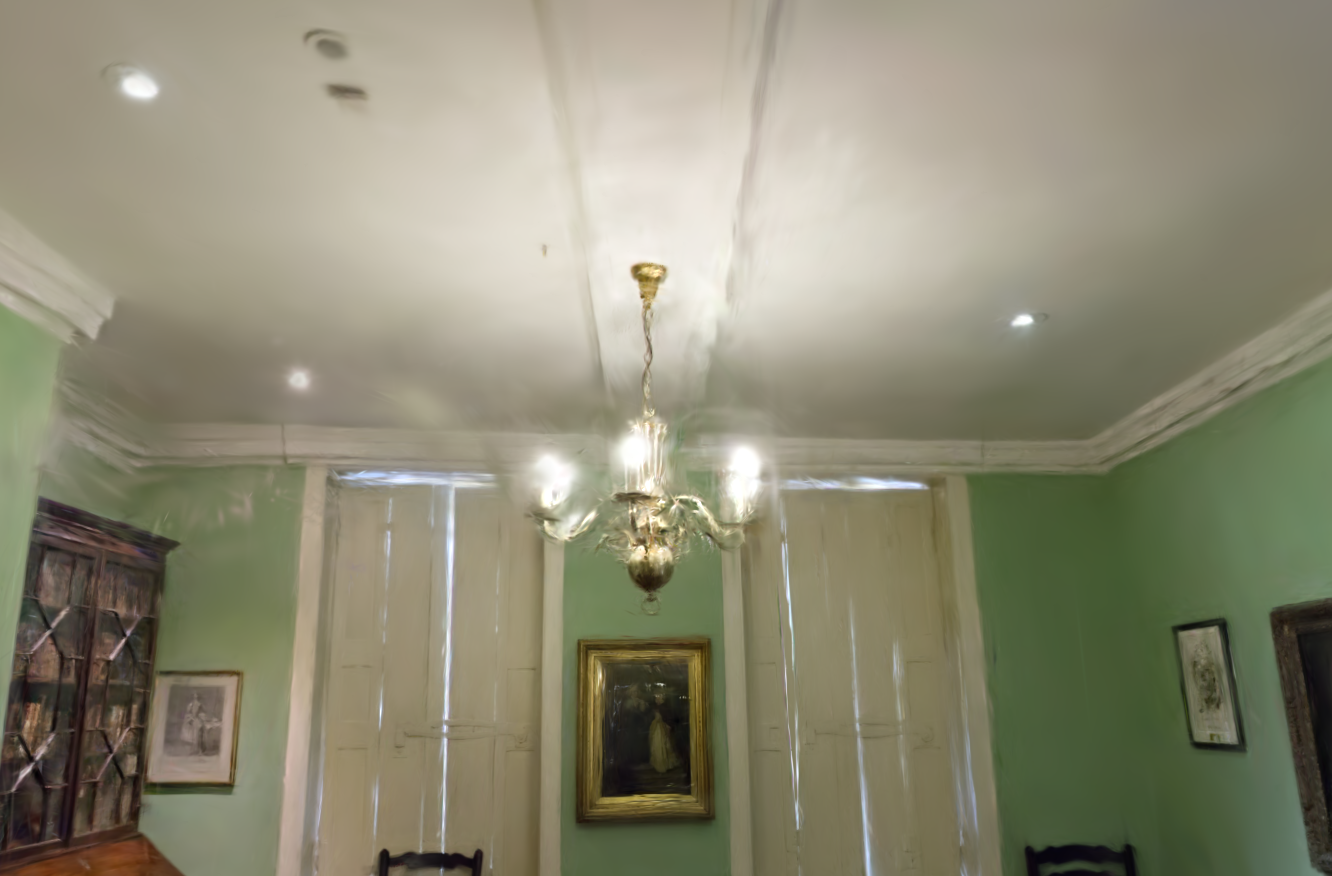} &
            \includegraphics[width=0.181\textwidth]{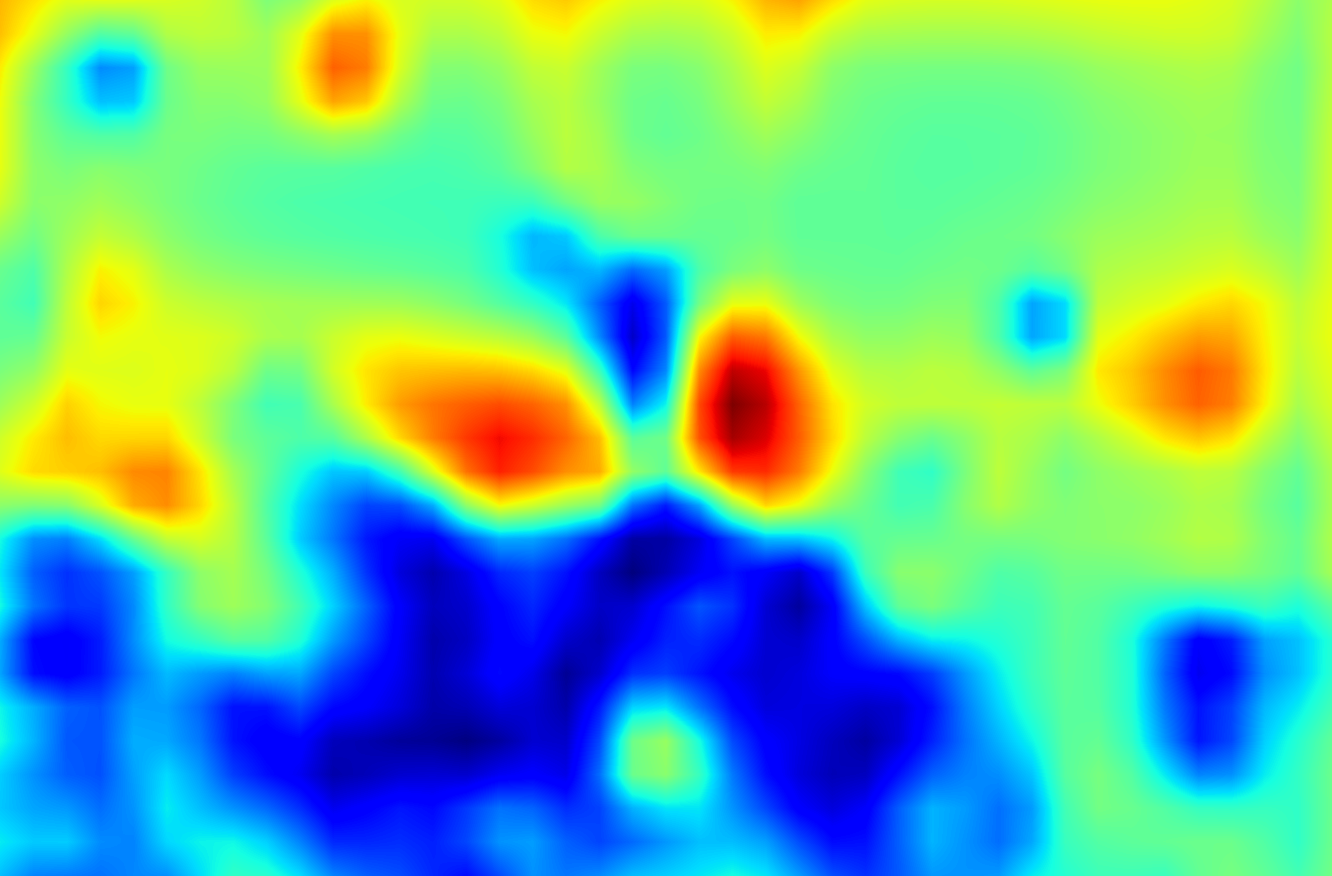} &
            \includegraphics[width=0.181\textwidth]{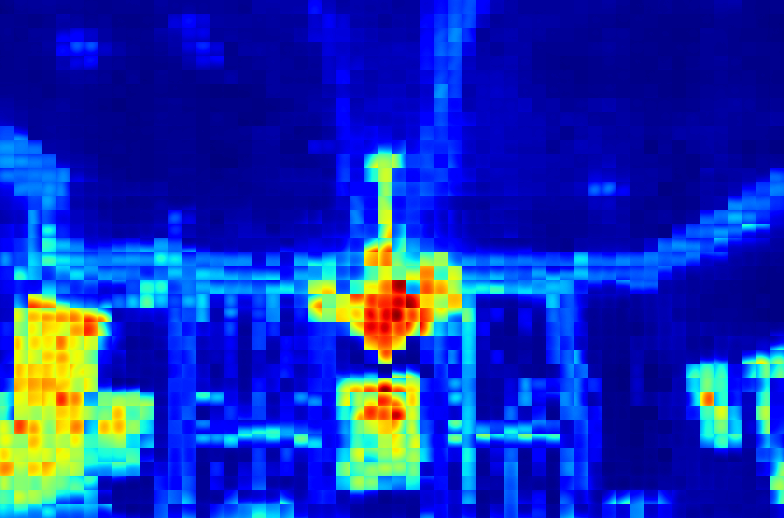} &
            \includegraphics[width=0.181\textwidth]{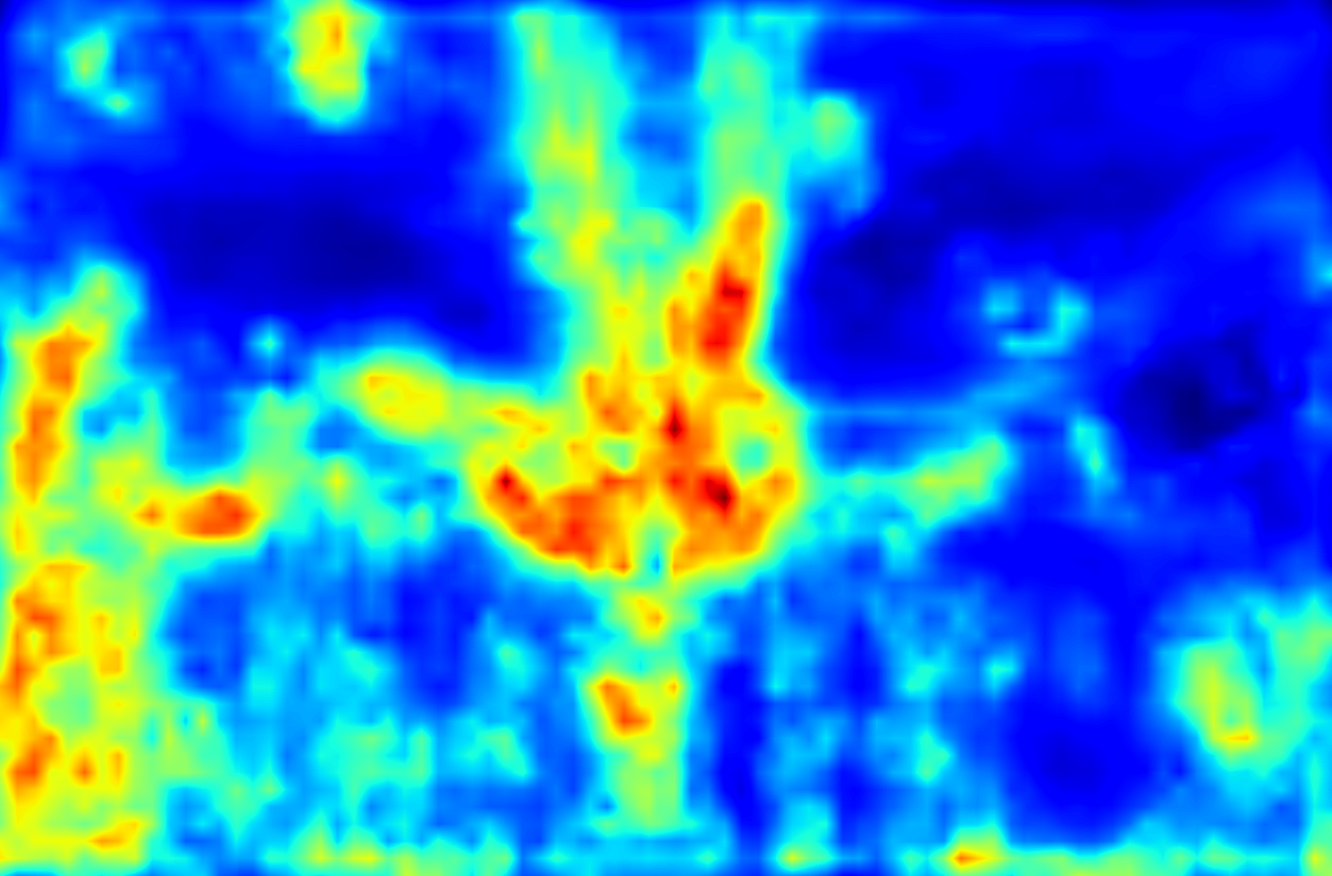} &
            \includegraphics[width=0.181\textwidth]{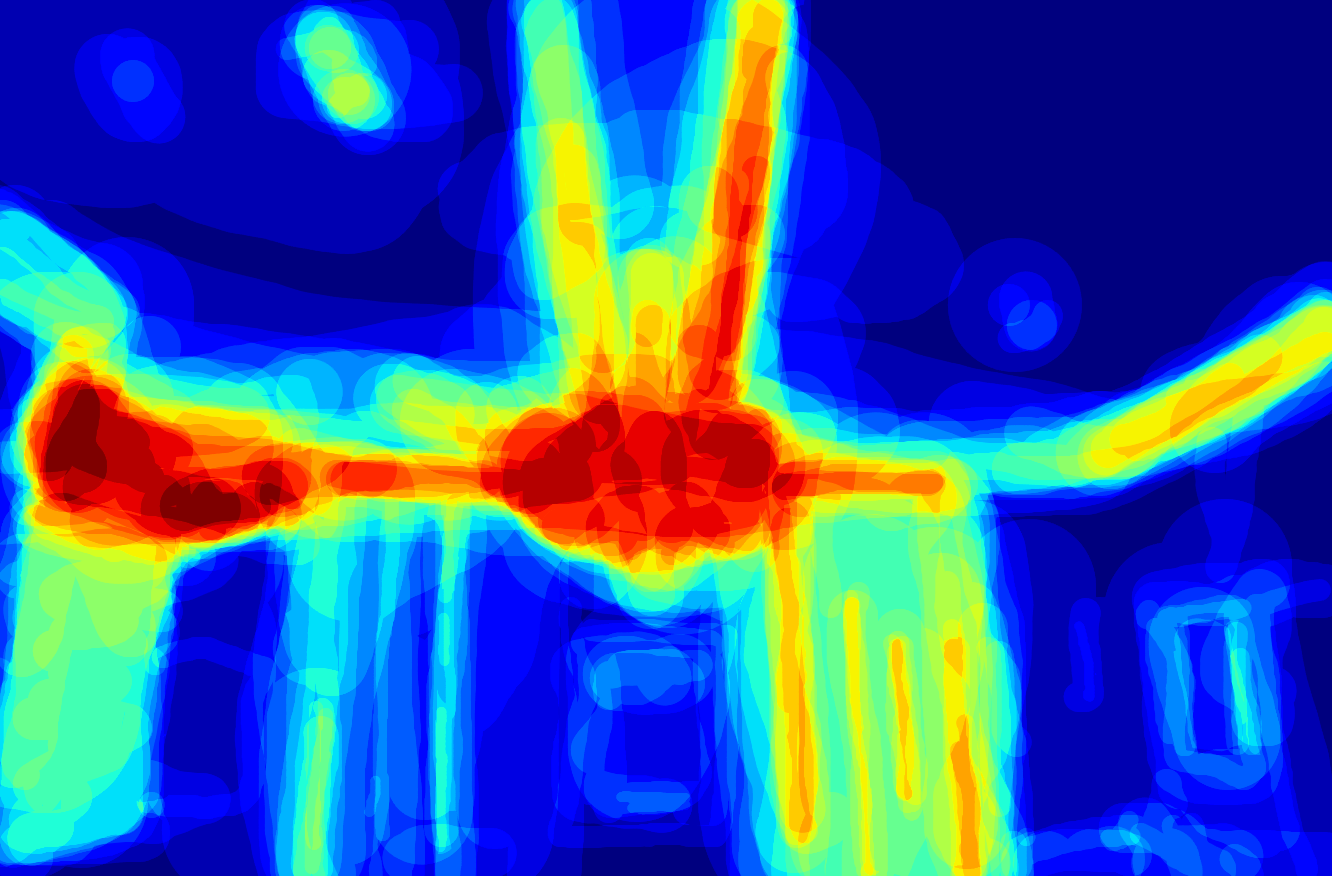} \\
            \includegraphics[width=0.181\textwidth]{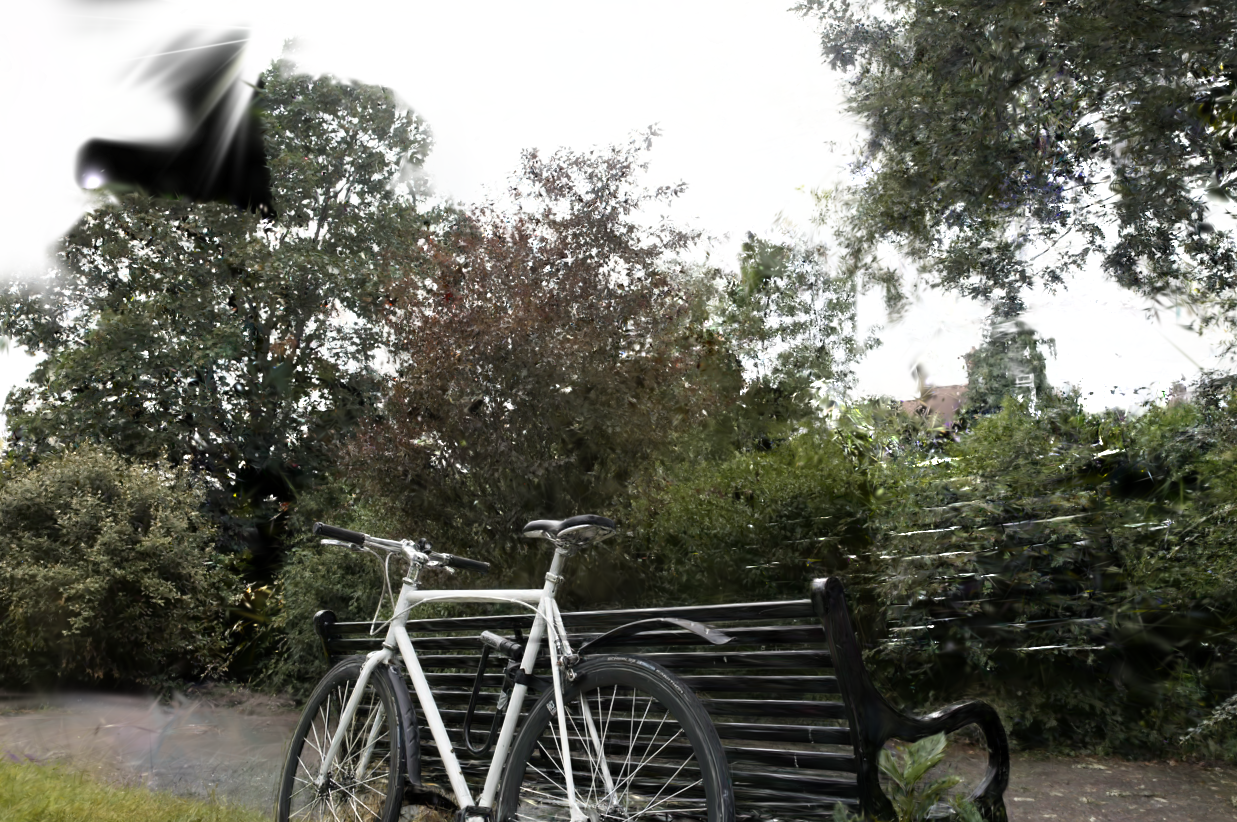} &
            \includegraphics[width=0.181\textwidth]{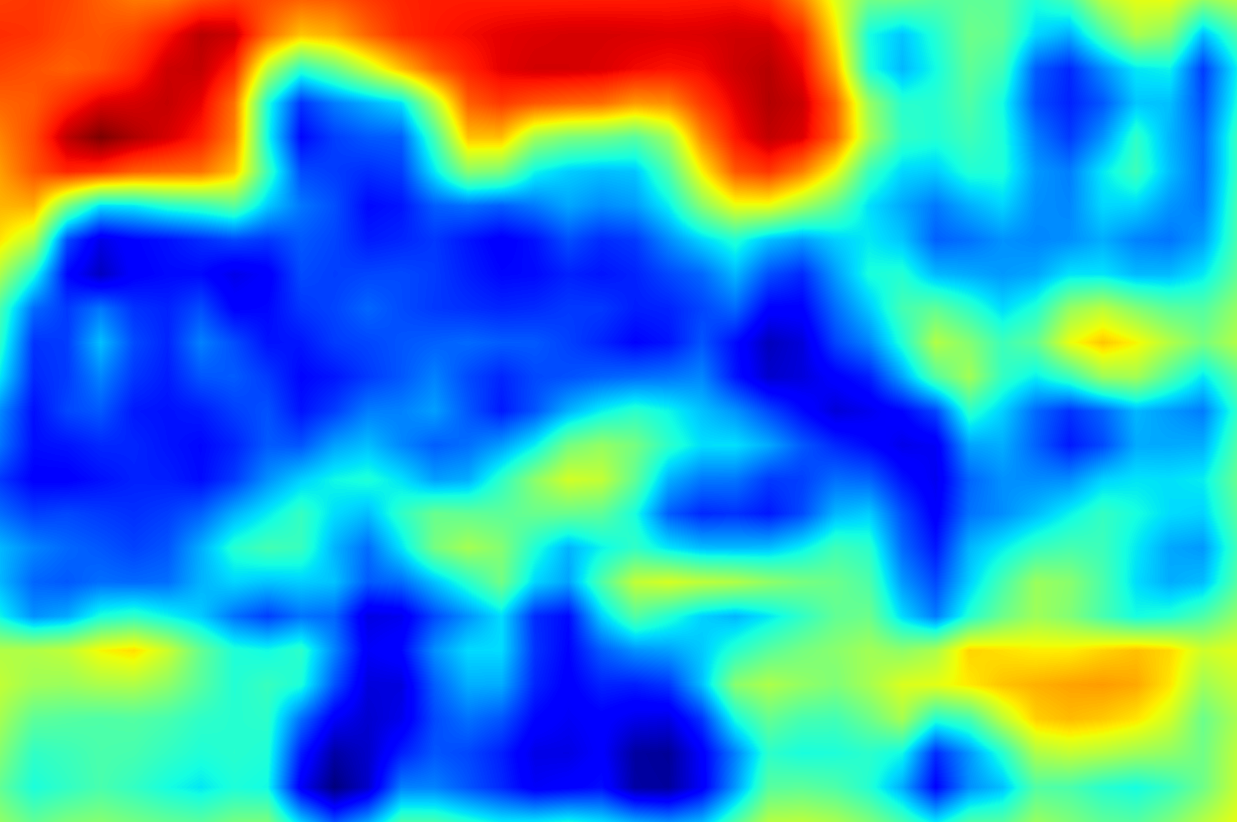} &
            \includegraphics[width=0.181\textwidth]{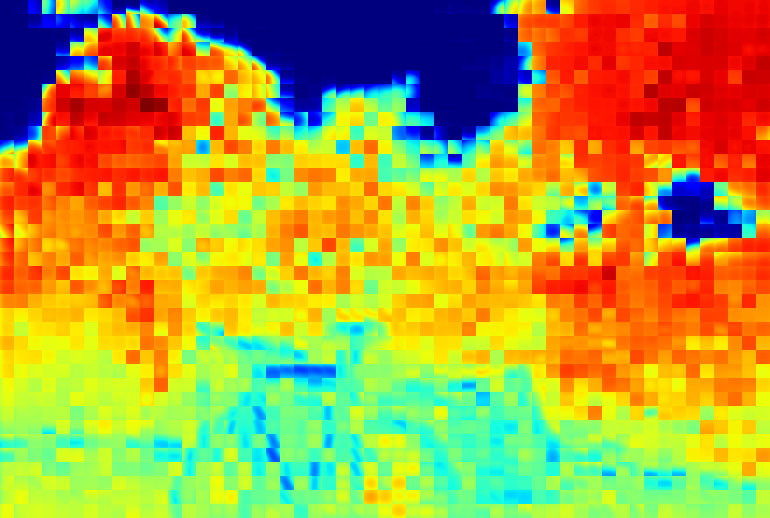} &
            \includegraphics[width=0.181\textwidth]{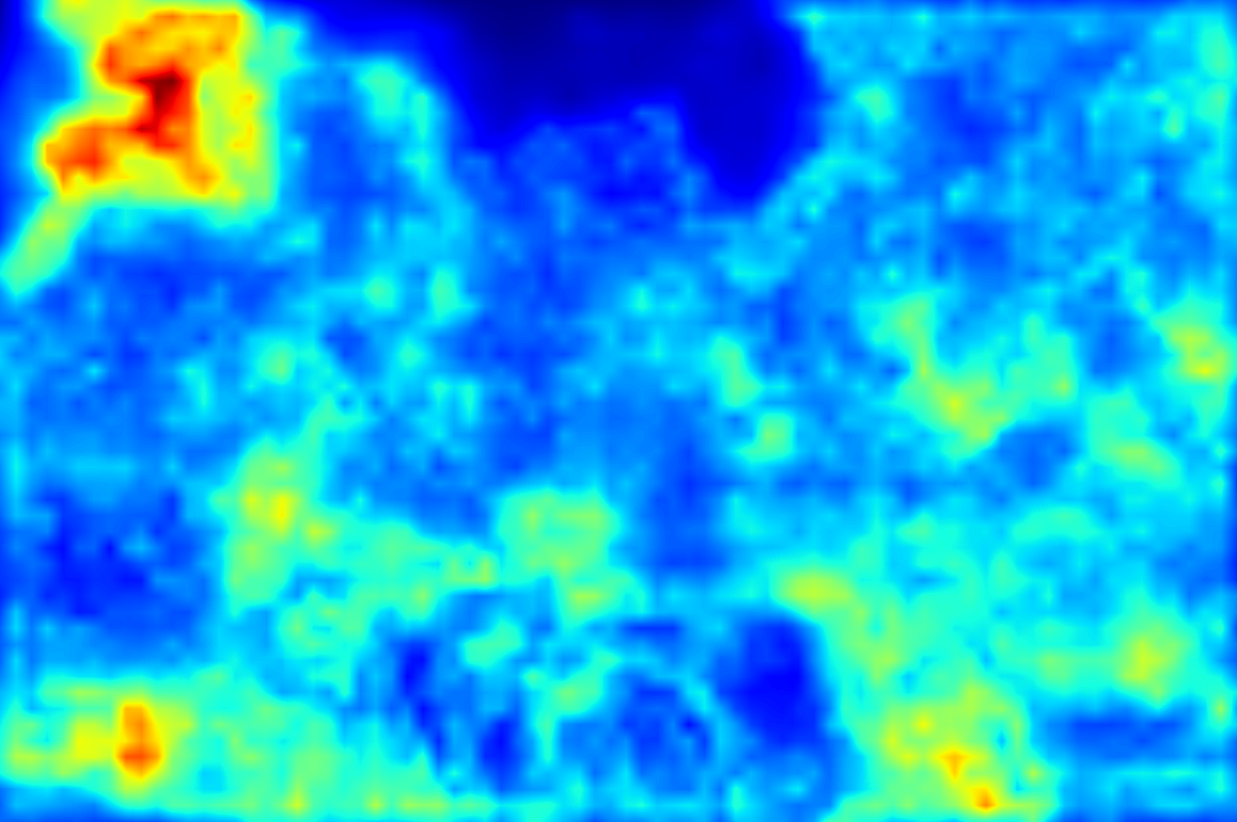} &
            \includegraphics[width=0.181\textwidth]{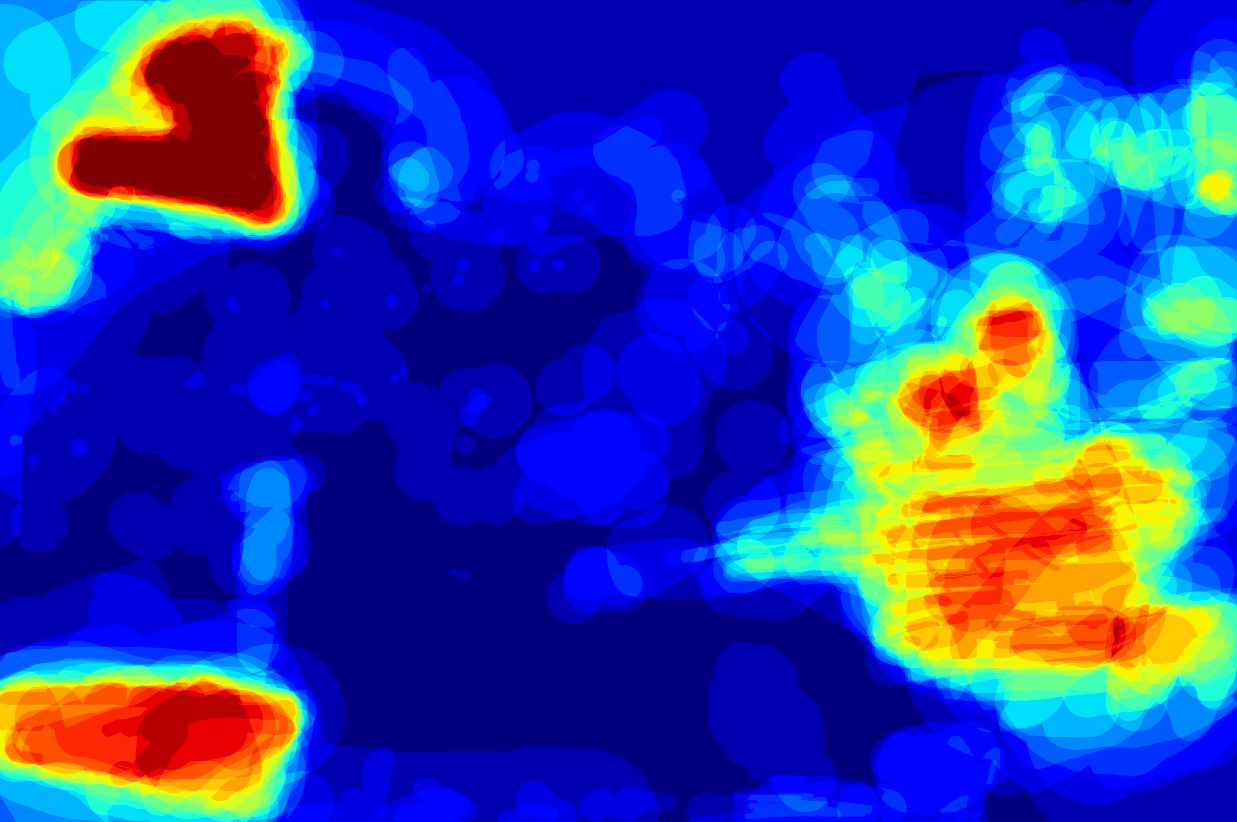} \\
            \subcaptionbox{Artifact-ridden view}[0.181\textwidth]{%
                \includegraphics[width=\linewidth]{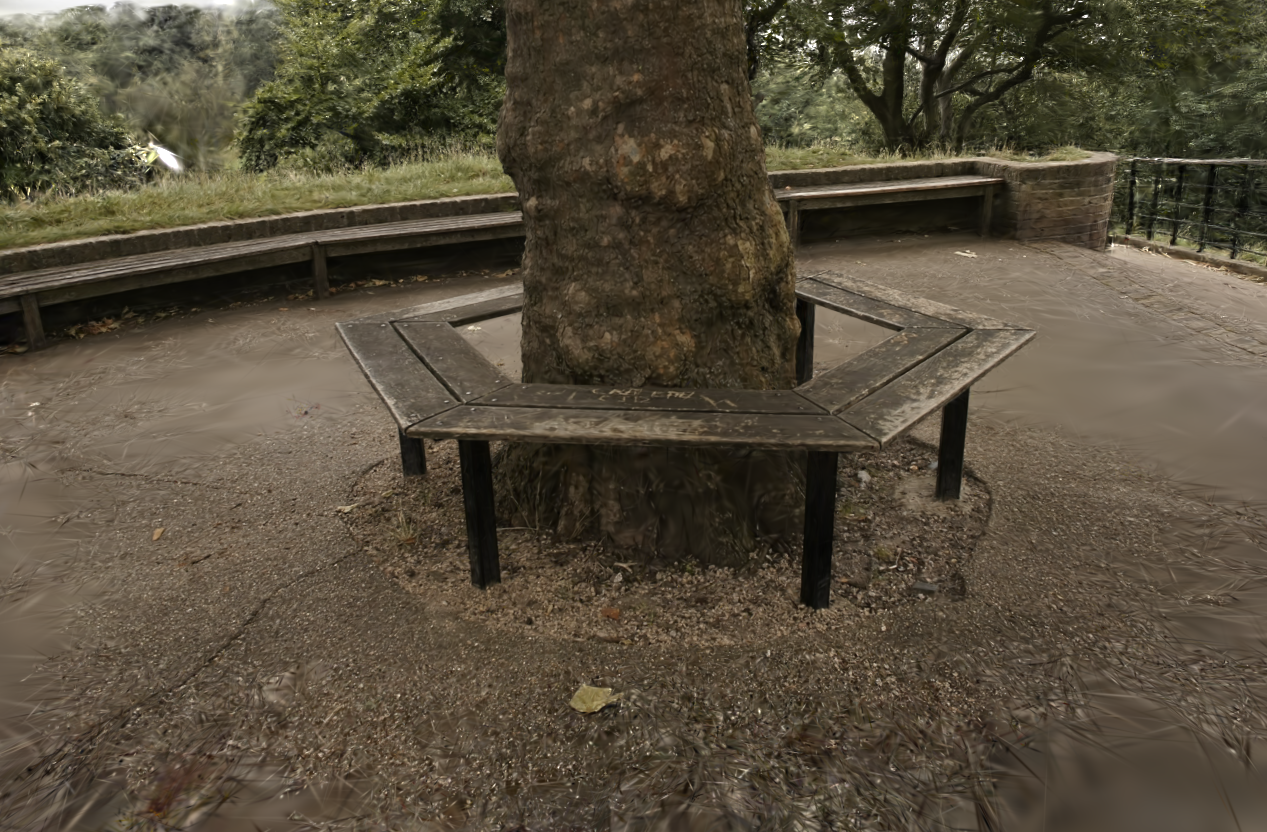}} &
            \subcaptionbox{PaQ-2-PiQ map}[0.181\textwidth]{%
                \includegraphics[width=\linewidth]{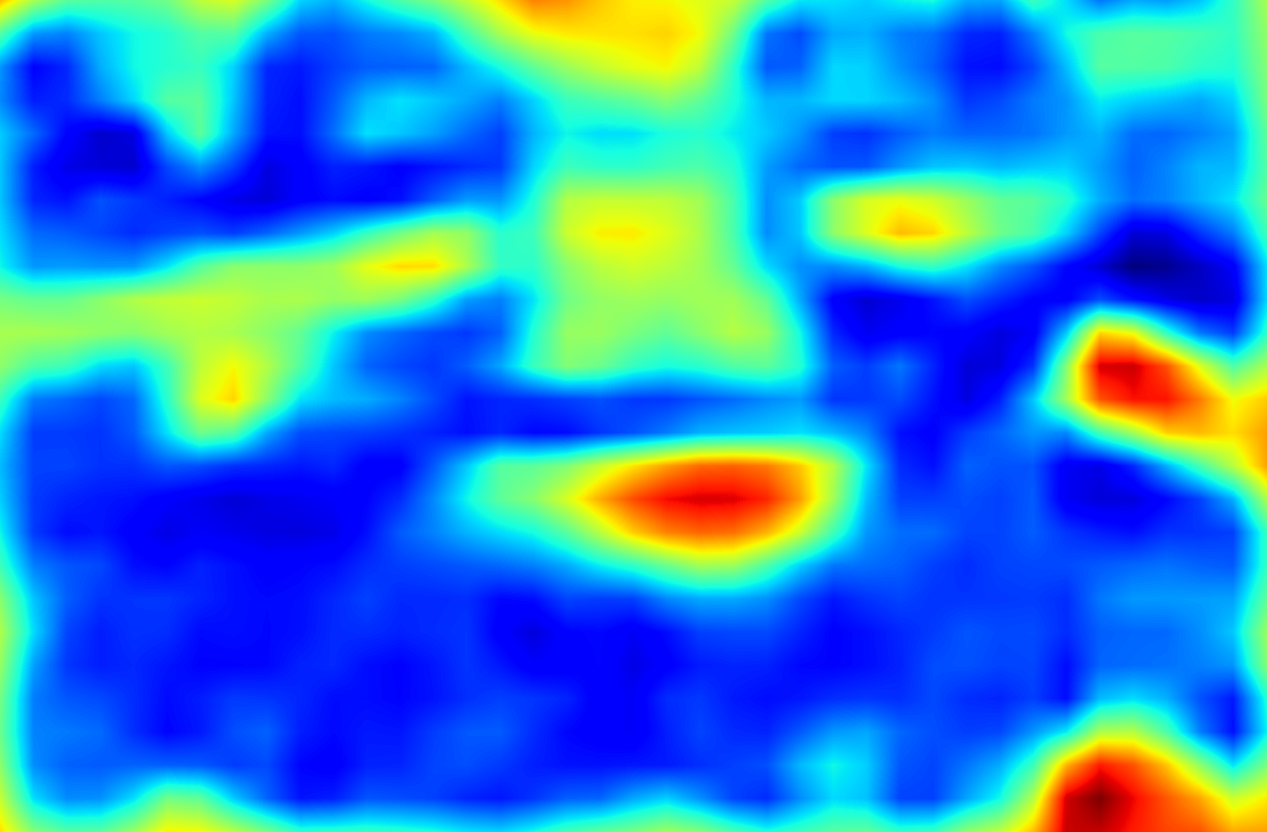}} &
            \subcaptionbox{CrossScore map}[0.181\textwidth]{%
                \includegraphics[width=\linewidth]{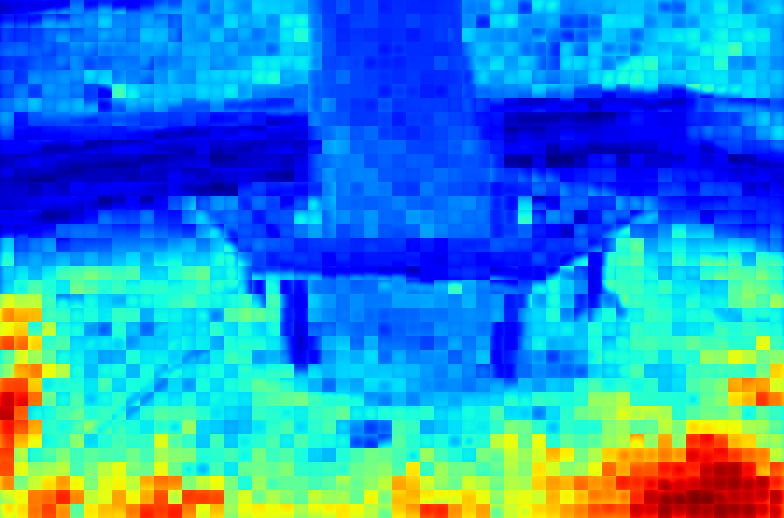}} &
            \subcaptionbox{PuzzleSim map (ours)}[0.181\textwidth]{%
                \includegraphics[width=\linewidth]{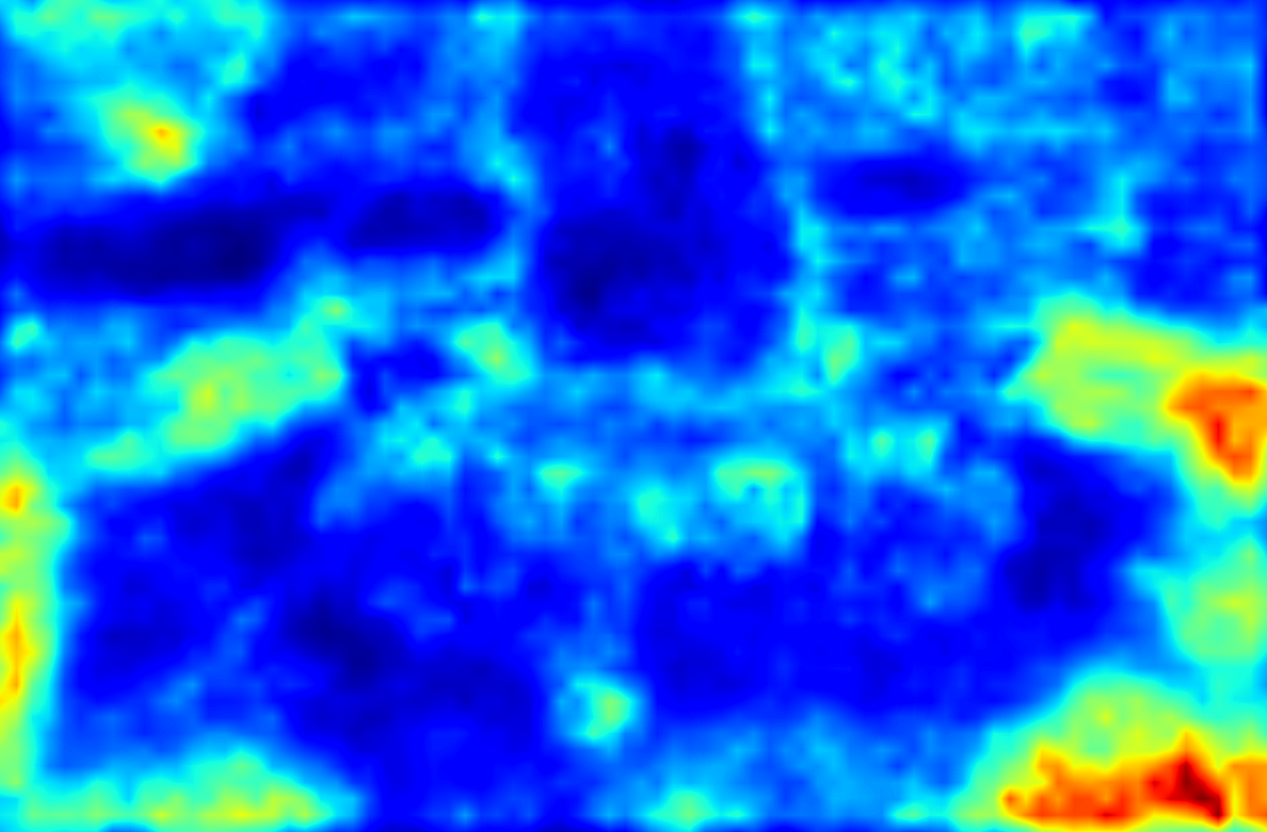}} &
            \subcaptionbox{Human labeling}[0.181\textwidth]{%
                \includegraphics[width=\linewidth]{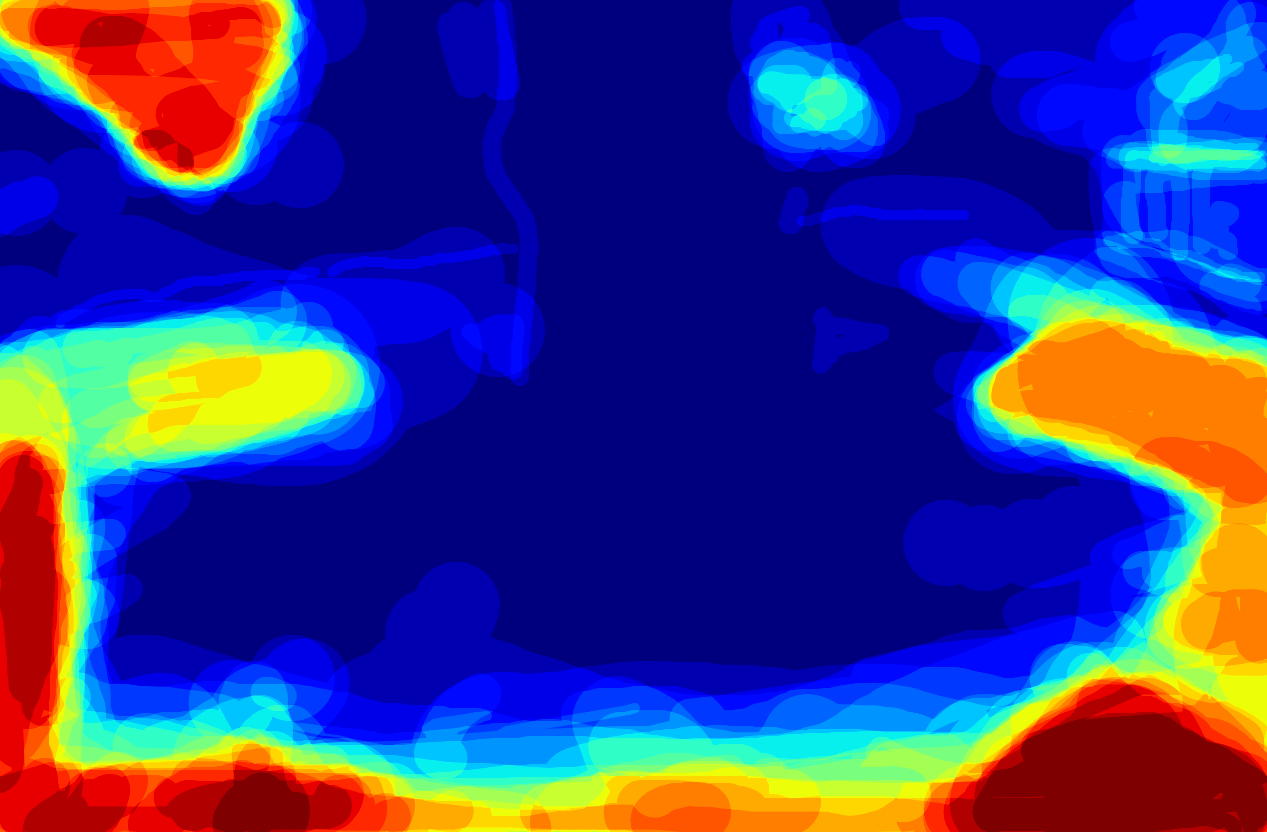}}
        \end{tabular}
    \end{adjustbox}
    \hspace{-0.015\textwidth}
    \begin{adjustbox}{valign=t}
        \begin{minipage}[t]{0.027\textwidth}
            \vspace{0pt}
            \includegraphics[width=1.176\textwidth]{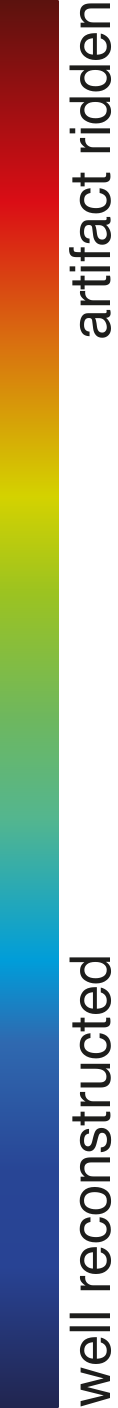}
        \end{minipage}
    \end{adjustbox}

    \caption{Selection of image quality maps for artifact-ridden renderings from various scenes. The last column shows ground-truth human assessments from our collected dataset. Note that our metric provides the finest resolution, enabling better artifact segregation.}
    \label{fig:examples-experiment}
\end{figure*}
}

%% file: tables/no_ref_table.tex
\begin{table*}
\caption{Pearson and Spearman Correlation between NR, and CR metrics and Human Perception per Dataset. The dashed line separates NR (above) from CR (below) metrics.}
\label{tab:nr_per_ds}
\small
\setlength{\tabcolsep}{4pt}
\begin{tabularx}{\textwidth}{llXXXXXXXXXXXX}
\toprule
 && \rotw{bicycle} & \rotw{bonsai} & \rotw{counter} & \rotw{drjohnson} & \rotw{flowers} & \rotw{garden} & \rotw{kitchen} & \rotw{playroom} & \rotw{stump} & \rotw{train} & \rotw{treehill} & \rotw{truck} \\
\midrule
\multirow{6}{*}{\rotatebox{90}{Pearson}}
&PAL4VST~\cite{zhang_perceptual_2023} & 0.139 & 0.088 & 0.062 & \30.153 & 0.002 & 0.005 & 0.068 & 0.197 & 0.000 & 0.104 & 0.000 & 0.119 \\
&PaQ-2-PiQ~\cite{ying_patches_2020} & 0.194 &\2 0.420 &\3 0.432 & 0.138 & 0.305 & 0.428 &\3 0.613 &\3 0.452 &\2 0.534 &\1 0.667 & 0.220 &\3 0.424 \\
&PIQE~\cite{n_blind_2015} & \30.266 & 0.267 & 0.255 & -0.089 &\2 0.583 &\3 0.441 & 0.490 & 0.091 &\3 0.526 & 0.200 &\2 0.376 & 0.101 \\
&CNNIQA~\cite{kang_convolutional_2014} & 0.027 & 0.037 & 0.064 & -0.068 & -0.053 & 0.409 & 0.324 & 0.345 & 0.367 & 0.400 & -0.133 & 0.005 \\
\cdashline{2-13}
&CrossScore~\cite{wang_crossscore_2025} &\2 0.338 &\3 0.331 &\2 0.493 &\2 0.390 &\3 0.476 &\1 0.748 &\2 0.663 &\2 0.603 &\1 0.585 &\3 0.565 &\3 0.300 &\1 0.630 \\
&\textbf{PuzzleSim (ours)} & \10.594 &\1 0.565 &\1 0.618 &\1 0.461 &\1 0.609 &\2 0.675 &\1 0.768 &\1 0.636 & 0.505 &\2 0.642 &\1 0.717 &\2 0.593 \\
\midrule\midrule
\multirow{6}{*}{\rotatebox{90}{Spearman}}
&PAL4VST~\cite{zhang_perceptual_2023} & 0.083 & 0.080 & 0.027 & 0.111 & 0.004 & -0.003 & 0.069 & 0.169 & 0.000 & 0.098 & 0.000 & 0.108 \\
&PaQ-2-PiQ~\cite{ying_patches_2020} &\3 0.214 &\1 0.495 &\1 0.435 & 0.009 & 0.261 & 0.176 &\3 0.616 & 0.329 &\2 0.476 &\1 0.696 & 0.152 &\3 0.329 \\
&PIQE~\cite{n_blind_2015} & 0.209 &\2 0.409 &\3 0.291 & -0.130 &\1 0.460 & 0.229 &\2 0.620 & 0.079 &\3 0.375 & 0.224 & \20.279 & 0.174 \\
&CNNIQA~\cite{kang_convolutional_2014} & -0.085 & 0.020 & 0.166 & \30.157 & -0.130 & \30.255 & 0.375 & \30.380 & 0.316 & 0.395 & -0.219 & -0.066 \\
\cdashline{2-13}
&CrossScore~\cite{wang_crossscore_2025} & \20.299 & 0.030 & 0.243 &\1 0.508 &\3 0.365 &\1 0.590 & 0.315 &\2 0.534 &\1 0.490 &\3 0.494 & \30.240 &\2 0.431 \\
&\textbf{PuzzleSim (ours)} &\1 0.468 &\3 0.393 &\2 0.382 &\2 0.499 &\2 0.428 &\2 0.428 &\1 0.658 &\1 0.601 & 0.307 &\2 0.540 &\1 0.548 &\1 0.440 \\

\bottomrule
\end{tabularx}
\end{table*}

%% file: tables/corr_new.tex
\begin{table}
\caption{Aggregated correlation between Image Metrics and Human Perception with mean and standard deviation across all datasets. Above the dashed line, we list NR, below CR metrics.}
\label{tab:corr}
\centering
\begin{tabularx}{\linewidth}{ll>{\centering \arraybackslash}X>{\centering \arraybackslash}X}
\toprule
\textbf{Metric} & Pearson $\uparrow$ & Spearman $\uparrow$ \\
\midrule
PAL4VST~\cite{zhang_perceptual_2023} & $0.078_{\pm0.112}$ & $0.062_{\pm0.085}$ \\
CNNIQA~\cite{kang_convolutional_2014} & $0.144_{\pm0.247}$ & $0.130_{\pm0.253}$ \\
PIQE~\cite{n_blind_2015} & $0.292_{\pm0.222}$ & $0.268_{\pm0.221}$ \\
PaQ-2-PiQ~\cite{ying_patches_2020} & \3$0.402_{\pm0.178}$ & \3$0.349_{\pm0.225}$ \\
\hdashline
CrossScore~\cite{wang_crossscore_2025} & \2$0.510_{\pm0.204}$ &\2 $0.378_{\pm0.209}$ \\
\textbf{PuzzleSim (ours)} & \1$0.615_{\pm0.120}$ & \1$0.474_{\pm0.137}$ \\

\bottomrule
\end{tabularx}
\end{table}

%% file: sec/5_application.tex
\section{Application: Progressive Inpainting} \label{sec:applications}
Finally, we showcase an application of our metric in automatic restoration of novel views from a reconstructed scene.
Whenever it is possible to establish a visual distribution (e.g., we have a training dataset available), we can recursively use our metric to automatically identify visual outliers in novel views and remove them through inpainting. \finalrev{In the Supplementary, we present a quantitative ablation study to show that our metric performs best in this application.}
\begin{figure*}[ht!]
    \centering
    \includegraphics[width=1\textwidth]{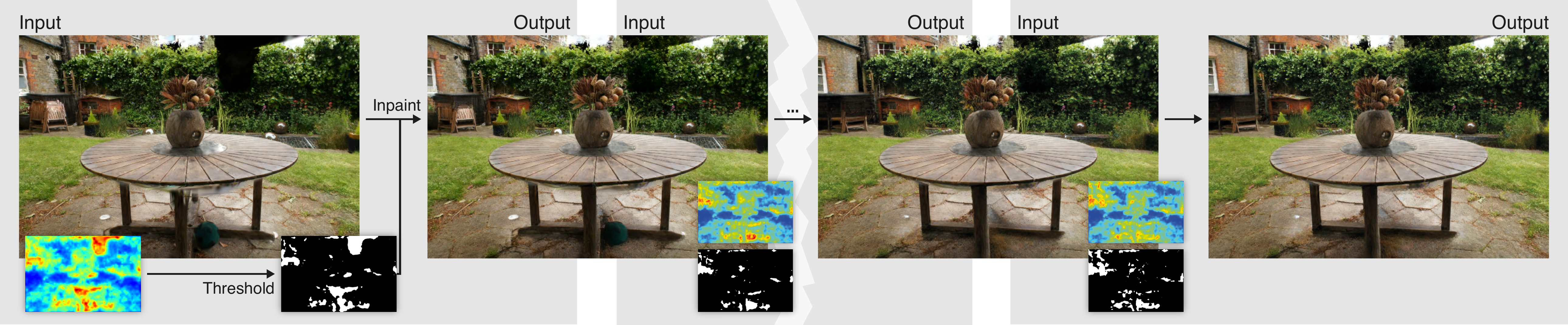}
    \vspace{.15cm}
    \includegraphics[width=1\textwidth]{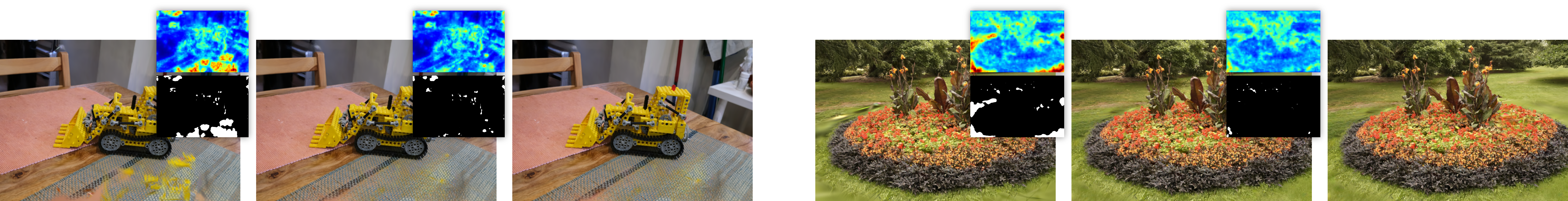}
    \caption{Example showcase of our iterative inpainting application to enhance new views that lack ground-truth correspondences.}
    \label{fig:inpainting_example}
\end{figure*}
\paragraph{Our Framework}
We can take a new image $\img$ and employ our \textit{PuzzleSim} metric to obtain the similarity map $\mathcal{S}$.
\begin{equation}
   \begin{aligned}
        \mathcal{S} =& \text{ PuzzleSim}(\img) \in \R^{H_\img \times W_\img}
    \end{aligned}
\end{equation}
To apply neural inpainting, we first need to create a binary mask from the similarity map $\mathcal{S}$, indicating the areas to be inpainted. This involves finding an optimal threshold $\tau$ that clearly distinguishes outlier regions. The effectiveness of inpainting depends on carefully setting this mask. If the mask is too large, the inpainting may inadvertently remove clean parts of the scene. If the mask is too small, artifacts might be left untouched. In order to automatically find a balanced threshold, we use a conservative, iterative approach to refine the test image based on the assumption that artifacts have below-average similarity scores. For an initial threshold, we select $N = 50$ candidate values, uniformly spaced between the lowest and mean similarity scores that we use to threshold the similarity map.
\begin{equation}
    \begin{aligned}\label{eq:initial_sampling}
        \tau_i =& \min(\mathcal{S}) + \frac{i}{N-1} (\text{mean}(\mathcal{S}) - \min(\mathcal{S})) \\ 
        M^{(h,w)}_i =& \begin{cases} 
            1 & \text{if } \mathcal{S}^{(h,w)} \leq \tau_i \\
            0 & \text{if } \mathcal{S}^{(h,w)} > \tau_i 
            \end{cases}
    \end{aligned}
\end{equation}
with $i$, $h$, and $w$ representing indices where $i = 0, \dots, N-1$, $h = 1, \dots, H_\img$, and $w = 1, \dots, W_{\img}$. With LaMa (big)~\cite{suvorov_resolution-robust_2022} we generate inpaintings using all $N$ masks and recompute \textit{PuzzleSim} for each option. The quality of each inpainted candidate is evaluated by calculating the average similarity difference before and after inpainting, denoted as $\delta_i$. To discourage overly large masks, we add a regularization term that penalizes them. Further details on the mathematical definitions of $\delta_i$ and the regularization term are provided in the Supplementary material. We then select the candidate that maximizes $\delta$. After determining the initial threshold, we iteratively refine the inpainted image by sampling new thresholds close to the previous one. The size of this interval depends on a hyperparameter $\alpha$ and the spread of similarity scores. Keeping this range small ensures stable convergence and prevents excessive, disrupting inpainting. If the upper limit of the interval is below the minimum similiarity value $\min_{h, w} \hat{\mathcal{S}}^{(h,w)}$, we revert to the initial sampling method in \cref{eq:initial_sampling} as an empty mask would be meaningless and cause division by zero when computing $\delta_i$. Finally, we terminate the process if no further improvement is achieved (i.e., $\max_i \delta_i \leq 0$), returning the final inpainting result. This framework guarantees a monotonic improvement in \textit{PuzzleSim} similarity.
In \cref{fig:inpainting_example}, we showcase several novel views from the reconstructed scenes \textit{garden}, \textit{kitchen}, and \textit{flowers} using only a fraction of the original training views (20-30\%). We process this artifact-ridden new view through the iterative inpainting framework presented above. Our method successfully detects and inpaints artifacts in the original reconstruction, producing high-quality inpainting consistent with the distribution of the original scenes. 

%% file: sec/6_limitations.tex
\section{Limitations and Future Work}\label{sec:limitations}
While our method demonstrates promising results, there are some limitations to consider. Even with our optimized implementation, finding the maximum similarity for a great number of vectors becomes expensive as the number of reference images and image resolution rise (see Supplemental material for implementation and runtime analysis). Performing approximate maximum search or fitting Gaussian mixture models in the embedding space can improve computational performance \cite{zhang_feature-enriched_2015, johnson_billion-scale_2021}. Furthermore, our metric is empirically calibrated, but choosing the weights to combine layers and weighting in the channel dimension in a data-driven manner could advance the metric further. The resolution at which our metric can be utilized is currently bound by the CNN backbone's generalizability to higher resolutions. \revision{}{Currently, we can not support Vision Transformer backbones as it takes special care to maintain a limited receptive field.} Although our metric is differentiable, it is unlikely to produce valuable gradients due to the max operation across many vectors. Alternatively, softmax operations could be explored to make the metric more suitable for gradient-based optimization.

%% file: sec/7_conclusion.tex
\section{Conclusion}\label{sec:conclusion}
We proposed Puzzle Similarity, a cross-reference image metric for detecting and localizing artifacts in novel views of 3D scene reconstructions. By leveraging learned patch statistics from input views, our method generates spatial artifact maps without requiring ground-truth references, addressing a key challenge in evaluating reconstructed scenes. To enable the evaluation of cross-reference metrics, we also provide a dataset of human-assessed quality and artifact localization for 3D scene reconstruction.

Our evaluation shows that Puzzle Similarity outperforms all tested full-reference, cross-reference and no-reference metrics in capturing artifacts aligned with human perception, demonstrating robustness across diverse artifact types and texture-rich scenes. Furthermore, we apply our metric to automatic image restoration, illustrating its potential to enhance scene reconstruction quality. Puzzle Similarity provides an effective, perceptually aligned, reference-free solution for artifact localization, with promising applications in few-shot reconstruction and guided acquisition.

\paragraph{Acknowledgements}
We would like to thank Krzysztof Wolski for making their image segmentation tool available to us, Volodymyr Kyrylov for providing the idea and first prototype of the memory-efficient implementation, and Sophie Kergaßner for designing figures. This project has received funding from the European Research Council (ERC) under the European Union’s Horizon 2020 research and innovation program (grant agreement N\textdegree 804226 PERDY), from the Swiss National Science Foundation (SNSF, Grant 200502) and an academic gift from Meta.